\documentclass[sigconf]{acmart}

\usepackage{booktabs} 
\usepackage{algorithm}
\usepackage[noend]{algorithmic}
\usepackage{graphicx}
\usepackage{url}

\setcopyright{none}

\renewcommand\footnotetextcopyrightpermission[1]{}
\begin{document}

\title{Finding Lookalike Customers for E-Commerce Marketing}

\author{Yang Peng}
\email{yang.peng@walmart.com}
\affiliation{Walmart Global Tech}

\author{Changzheng Liu}
\email{changzheng.liu@walmart.com}
\affiliation{Walmart Global Tech}

\author{Wei Shen}
\email{wei.shen@walmart.com}
\affiliation{Walmart Global Tech}


\begin{abstract}

Customer-centric marketing campaigns generate a large portion of e-commerce website traffic for Walmart.
As the scale of customer data grows larger, expanding the marketing audience to reach more customers is becoming more critical for e-commerce companies to drive business growth and bring more value to customers.
In this paper, we present a scalable and efficient system to expand targeted audience of marketing campaigns, which can handle hundreds of millions of customers.
We use a deep learning based embedding model to represent customers and an approximate nearest neighbor search method to quickly find lookalike customers of interest.
The model can deal with various business interests by constructing interpretable and meaningful customer similarity metrics.
We conduct extensive experiments to demonstrate the great performance of our system and customer embedding model.

\end{abstract}


\maketitle

\section{Introduction}

In customer relationship management (CRM) systems, customer acquiring and retention are crucial for marketing success.
Expanding the set of targeted customers is a very important component in CRM systems for both customer acquiring and retention.
In this article, we consider the problem of building a large scale marketing audience expansion system, aiming at driving e-commerce growth by finding more customers for CRM email and push marketing campaigns.

Formally speaking, the problem to solve in this paper is: given a set of existing customers (seed customer set) for a marketing campaign, 
how to find more customers that are similar to these seed customers, 
so that we can increase revenue and drive more traffic for Walmart e-commerce.

One of the biggest challenges we face in this problem is the scale of the data. 
Walmart has hundreds of millions of active customers in the US market alone.
Each customer could have hundreds or even thousands of features.
And customer data is increasing rapidly year by year.
Thus building a scalable and efficient system to handle ever-increasing big customer data is our top priority.
Besides scalability, we need to take into account the interpretability of our method when finding lookalike customers.
Model interpretability not only helps explain how our method works to business partners, but also provides an intuitive way for examining system quality.

In this article, we propose a scalable and efficient audience expansion system for marketing campaigns, which can handle hundreds of millions of customers.
\begin{itemize}
\item Our system can generate low dimensional dense embeddings to represent customers.
\item In the customer embedding space, we use the cosine distance between the two customer embedding vectors as the estimation of the similarity between two customers.
\item The similarity metric we design can measure different business interests (such as purchases, visits and engagements), which are interpretable and meaningful business goals for marketing campaigns.
\item We use approximate nearest neighbor search to quickly find lookalike customers to the seed customers in the customer embedding space.
\end{itemize}

To improve the quality of the customer embedding model, our model ingests multimodal features from different data sources, such as transactions, visits, engagements and customer metadata. 
Multimodal fusion techniques have demonstrated great benefits for tasks such as information retrieval, information extraction and classification \cite{nia2014streaming,niauniversity,peng2015probabilistic,peng2016scalable,peng2016multimodal,peng2017multimodal, acm2017web,peng2022qa,peng2022kbc,peng2023web} by leveraging the complementary and correlative relations between different types of data.
Our embedding model can also achieve better quality for audience expansion by combining different types of data.

Our contributions are shown below:
\begin{itemize}
	\item We propose an effective and efficient marketing audience expansion system, which can handle hundreds of millions of customers. 
	\item We use a deep learning model to generate customer embeddings. The deep learning model can handle both dense numerical features and sparse categorical features. And the model encodes location embedding using transfer learning.
	\item Our system has great adaptability by constructing different similarity metrics for different campaigns and business goals. The similarity metrics are both interpretable and meaningful.
	\item We develop a scalable and efficient approximate nearest neighbor search method based on FAISS to quickly find similar customers. 
	\item 
	We design multimodal features from various data sources.
	\item 
	Extensive experiments have been conducted to demonstrate the scalability, efficiency and quality of our system and embedding model.
\end{itemize}

\noindent \textbf{Overview} 
Related work on lookalike modeling and audience expansion systems is discussed in Section 2.
The overview of our system is presented in Section 3.
The embedding model is illustrated in Section 4.
We demonstrate the effectiveness and efficiency of our system through extensive experiments in Section 5.
The conclusions and future work of our lookalike model are discussed in Section 6.
\section{Related Work}

In this section, we will discuss the previous work on audience expansion systems, use representation models and fast similarity search methods.
For the literature review, we mostly focus on related work in industry solutions, since we are tacking large-scale or even web-scale datasets which are very rarely studied in academic research. 

\begin{figure*}[t]
	\centering
	\includegraphics[scale=0.8]{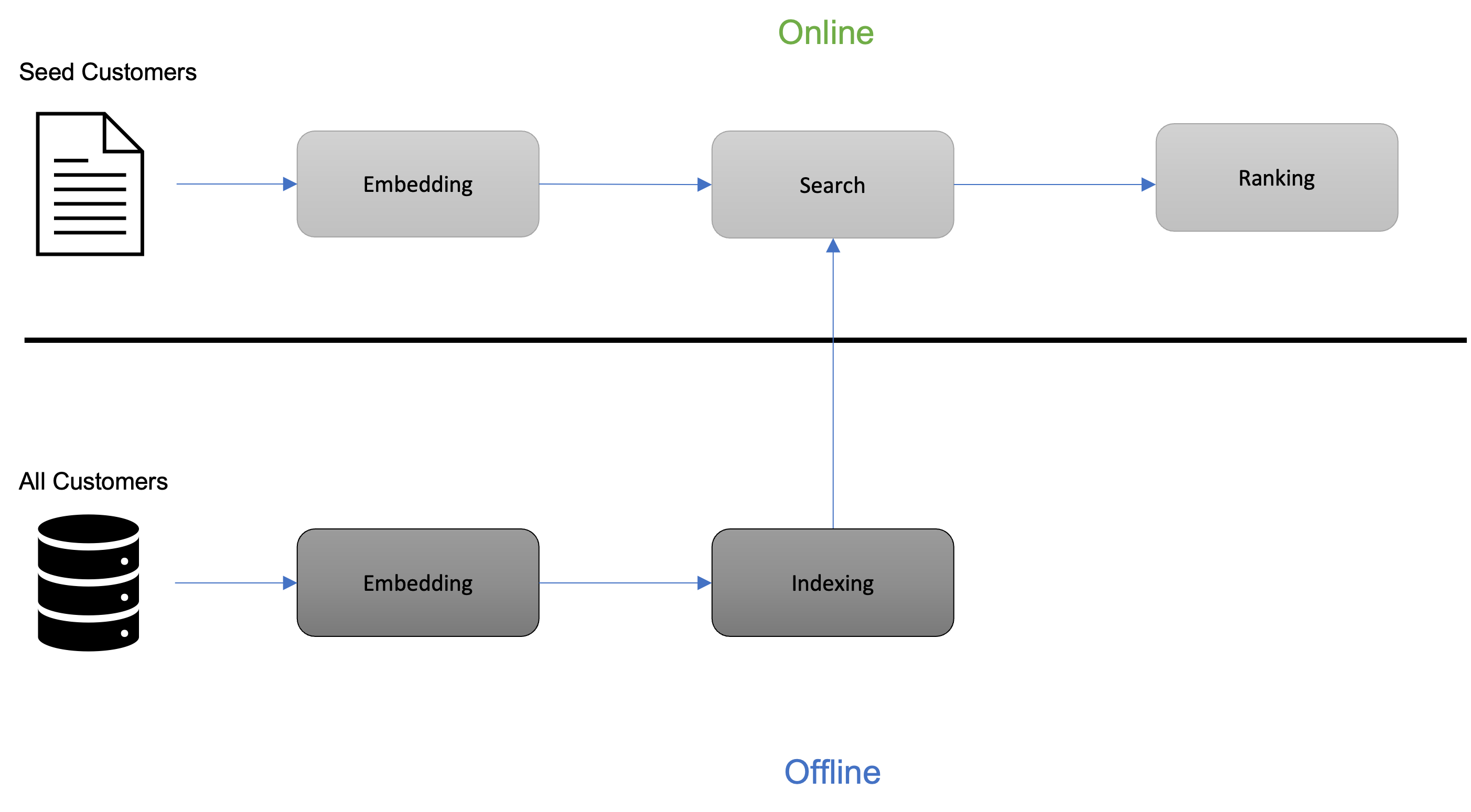}
	\caption{Audience Expansion System in Walmart.} \label{system}
\end{figure*}

\subsection{Audience Expansion}

In \cite{pmlrma16}, Ma et.al. discussed three types of approaches for audience expansion for marketing campaigns: similarity-based, regression-based and segmentation-based approaches. They proposed a graph-based lookalike system in Yahoo! advertising platform, which takes advantages of both simple similarity and regression-based methods. 
In \cite{liu2016audience}, Liu et.al. developed two methods to achieve audience expansion in LinkedIn: campaign-agnostic expansion based on user attributes and campaign-aware expansion using nearest neighbor search.
In \cite{jiang2019comprehensive}, Jiang et.al. discussed rule-based, similarity-based and model-based methods for finding lookalike users and proposed a deep neural network classification model for audience expansion in MiningLamp.
In \cite{dewet2019finding}, deWet et.al. proposed a two-stage embedding-based audience
expansion model that is deployed in production at Pinterest. For the first stage, they trained a global user embedding model on sitewide user activity logs. In the second stage, they used statistical techniques to create lightweight seed list representations in the embedding space for each advertiser. 

In our system, we first use a similarity-based approach to search lookalike customers from the whole customer universe and then rank these new customers based on their similarity scores or a separate classification model.
We choose the similarity-based approach for the first step because of its great scalability and low search latency. 
Similarity-based approaches usually require building user representations first.
In the next section, we will discuss embedding models for representing customers.

\subsection{User Representation}
Embedding models are very useful in terms of transforming high dimensional sparse feature vectors of customers to low dimensional dense representations of customers.
Embedding models have been widely used in industry, for example, for search engine marketing at Walmart \cite{jie2021bidding,jiemulti2022,jiedeep2022}, search ranking at Airbnb \cite{grbovic2018real}, and recommendation at Pinterest \cite{pal2020pinnersage}.
Embedding models can be trained in a way to capture similarity between customers, so that we can use approximate nearest neighbor search methods to quickly find lookalike customers of seed customers.
In our case, we use a two-tower architecture to train the customer embedding model, which is well recognized in previous work.
Our novelty in user representation is modeling business metrics using the cosine similarity between two embedding vectors, which has great interpretability.

\subsection{Similarity Search}
After user representation, we need a fast approach to search lookalike customers from a very large customer universe with potential size of hundreds of millions of customers. 
Scanning the whole customer universe is not scalable or efficient \cite{zhang2016construction,amirrahmat2018,Senseney2017,luo2022constitutive,Zhang2015fracture}.
Approximate nearest neighbor search is the most popular approach used in previous work.
For example, locality sensitive hashing (LSH) has been used in previous work \cite{pmlrma16,liu2016audience,dewet2019finding}.
There are several good open-source tools \cite{zhu2020high, jarrow2021low, zhu2021time, zhu2020adaptive, zhu2021news, zhu2021clustering, li2021frequentnet} for approximate nearest neighbor search, such as ScaNN \cite{scann2020} by Google and FAISS \cite{faiss2019} by Facebook. 
We choose FAISS for lookalike customer search for its scalability of handling billions of vectors and support of various types of distance measures (e.g. dot-product, cosine, Euclidean distances).

\section{System Overview}

In this section, we explain our audience expansion system pipeline for finding lookalike customers for e-commerce marketing in Walmart.
The system diagram is shown in Figure~\ref{system}.
Our audience expansion system has an online stage and an offline stage.
In the offline stage, we generate customer embeddings for all the customers in the customer universe and then build indexing on the customer embedding space. 
In the online stage, the seed customers are transformed into embeddings and then we search for lookalike customers using the pre-built indexes from the offline stage.
After getting the lookalike customers, we will conduct filtering and ranking on them.
The ranking approach can be based on their similarity scores or a different classification model.

\subsection{Embedding}

In this article, we will mostly focus on the embedding model and explain how we build this model in later sections.
The customer embedding model yields unified dense representations of customers.
Our customer embedding model can be utilized in a lot of use cases.
Besides finding lookalike customers, customer embeddings can be employed as input features in other customer models, such as purchase propensity models and life-time value models.
Ranking method is not a focus in this paper and will be studied in our future work.

\subsection{Indexing and Search}
The scalability issue in this system is how to quickly search for lookalike customers from a pool of hundreds of millions of candidate customers.
To narrow down the customers for consideration, we build indexes using FAISS \cite{faiss2019}.
We choose FAISS for a few reasons as listed below.
FAISS supports a wide range of different indexes and provides both CPU and GPU implementations.
FAISS also supports different types of distance measures, such as Euclidean distances, dot products and cosine similarity.
And we can utilize compression techniques in FAISS to process large datasets that cannot fit in memory.
How we implement indexing and search is not the main focus of this paper. 
If you are interested in more details about FAISS, please visit their Github project page.

\section{Embedding Model}

In this section, we present our deep learning based embedding model, which can capture the similarity between customers.
To design this model, we need to first define the similarity metric between two customers.
While some models in previous work \cite{dewet2019finding} learned relative similarity scores between positive and negative customer pairs, we use direct similarity metrics between two customers, which are interpretable, meaningful and reflecting business metrics.
The similarity metrics we define can be very useful in improving and explaining marketing campaign performance.

We use a two-tower architecture to calculate the cosine distance between two customer embedding vectors and use the cosine distance as the estimate of similarity metric between two customers.
The customer embedding model is trained to minimize the total loss between cosine distances and true similarity scores in the training datasets.

\subsection{Similarity Metric Definition}

In Walmart, we care about a lot of different business metrics for marketing campaigns, such as transactions, website visits, campaign engagements.
When expanding audience for marketing campaigns, it's a business advantage to find new customers that have similar behavior on Walmart e-commerce website as the existing seed customers.
Finding new customers with similar business metrics can allow marketing campaigns to maintain a similar customer distribution after expansion, which is very beneficial for cold-start campaigns or conversion campaigns.

There are three types of business metrics of particular interest to our marketing campaigns: transactions, visits and engagements.
Let's take transactions as an example to illustrate how we define the similarity metrics.
Let's say there are a list of product categories in Walmart catalog, $c_1, c_2, ..., c_n$.
Customer A has made purchase orders in these categories, $O_A = (a_1, a_2, ..., a_n)$.
Customer B has also made purchase orders in these categories, $O_B = (b_1, b_2, ..., b_n)$.
The similarity between A and B is defined as:
\begin{equation}
	\label{sim}
	similarity(A, B) = cosine\_similarity(O_A, O_B)
\end{equation}
We can also use other similarity distance functions, such as Jaccard similarity and Euclidean distance.
This method of calculating similarity metrics can also be applied for visits and engagements.

\begin{figure}[h]
	\centering
	\includegraphics[scale=0.6]{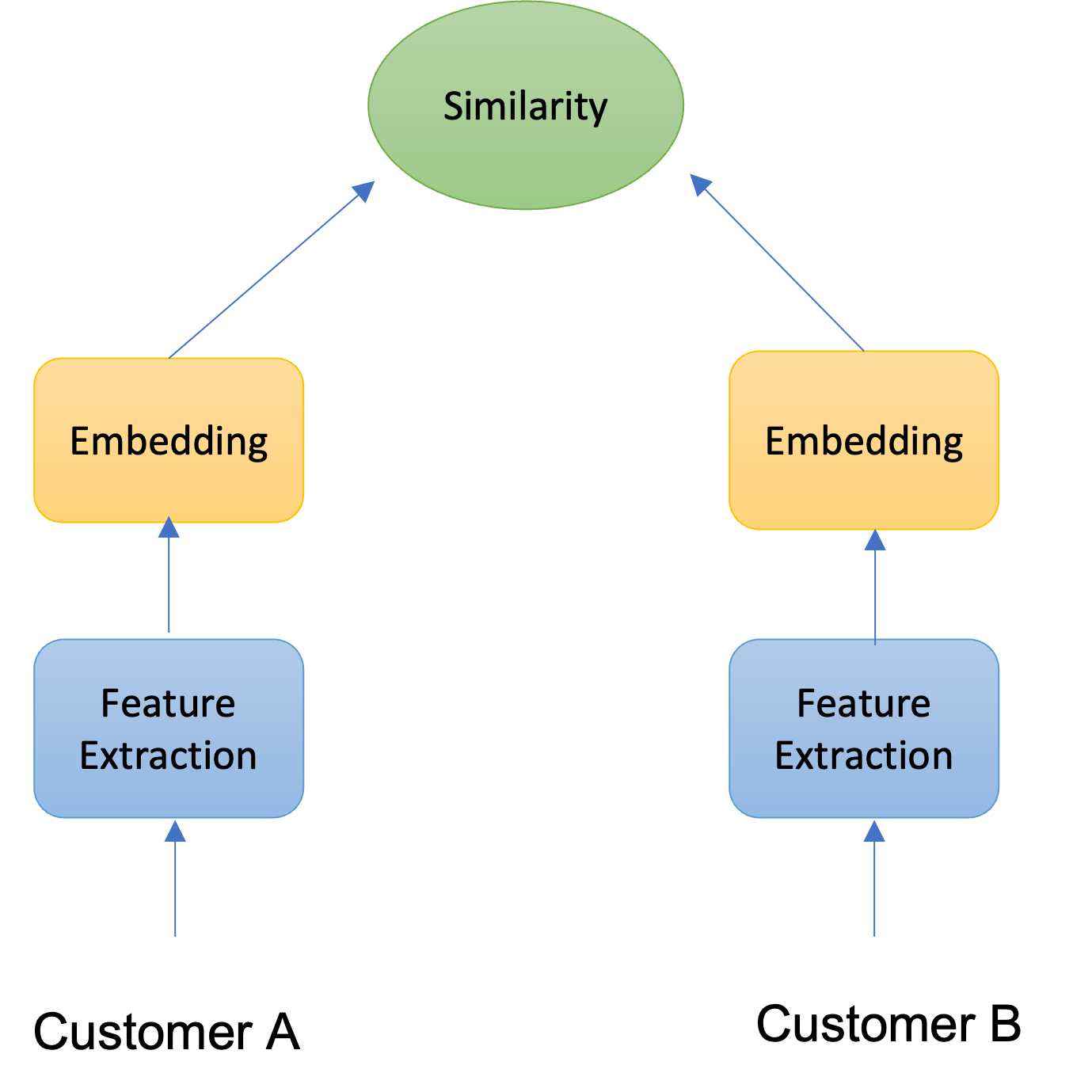}
	\caption{Two-Tower Model Architecture.} \label{tower}
\end{figure}

\subsection{Two-Tower Architecture}
\begin{figure*}[t]
	\centering
	\includegraphics[scale=0.7]{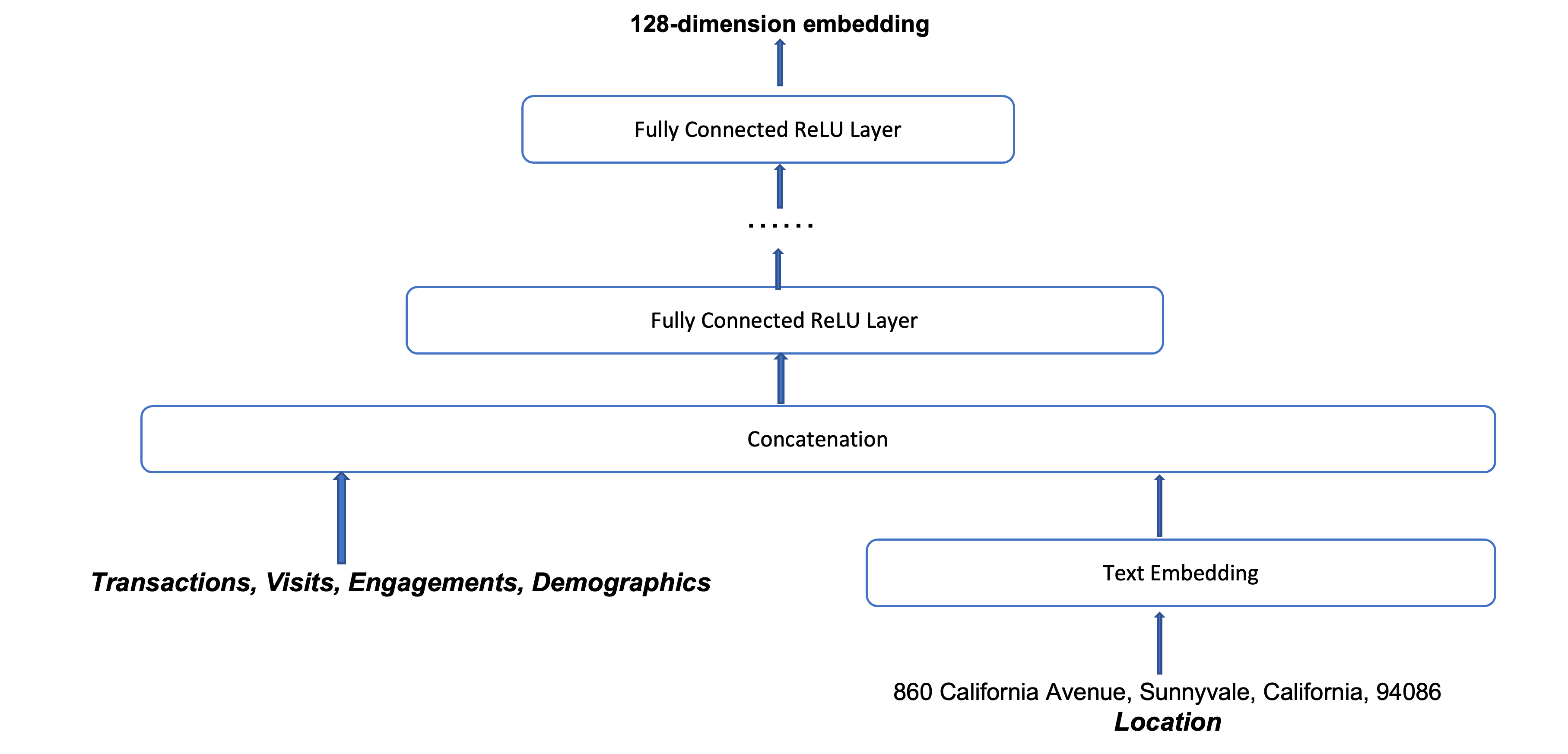}
	\caption{The Embedding Model.} \label{embedding}
\end{figure*}
After defining the similarity metric between customers, the next task is to build a machine learning model to predict the similarity metric given customer features.
Our approach is:
\begin{enumerate}
    \item extract raw features for customers;
    \item transform customer features into customer embeddings;
    \item calculate the cosine distances of customer embedding pairs in the embedding space;
	 \item use the cosine distances as the estimates of true similarity scores of customer pairs.
\end{enumerate}
The process is shown in Figure \ref{tower}.
The loss function for optimization is the L1 loss between cosine distance prediction and true similarity metric.

The multimodal customer features are extracted from multiple data sources, including transaction data, visit data, engagement data and demographics data.
The multimodal customer features are composed of dense numerical features (such as number of orders, GMV, number of visits) and sparse categorical features (such as gender, education, occupation, location).
For low dimensional categorical features, we can use one-hot encoding to transform them into numbers, for example gender and education level.
The location feature contains street address, city name, state name and zip code, so it's a very high dimensional categorical feature, which is too inefficient to use one-hot encoding.
We convert the location feature into location embedding using transfer learning and then concatenate it with other features together, which is explained in the next part.

\subsection{Embedding Model Structure}

The embedding model is described in Figure \ref{embedding}.
We have the transaction, visit, engagement, demographics and location features as input to the embedding model.
The location features are treated as textual sentences and then transformed to location embeddings using transfer learning of
pre-trained text embedding models.
Then we concatenate the numerical features and location embeddings as input to the final feed-forward neural network. 
The feed-forward network is composed of multiple fully connected ReLU layers.
The output of the feed-forward network is a 128 dimensional embedding as customer representation.

\subsubsection{Location Embedding}

\begin{figure}[h]
	\centering
	\includegraphics[scale=0.5]{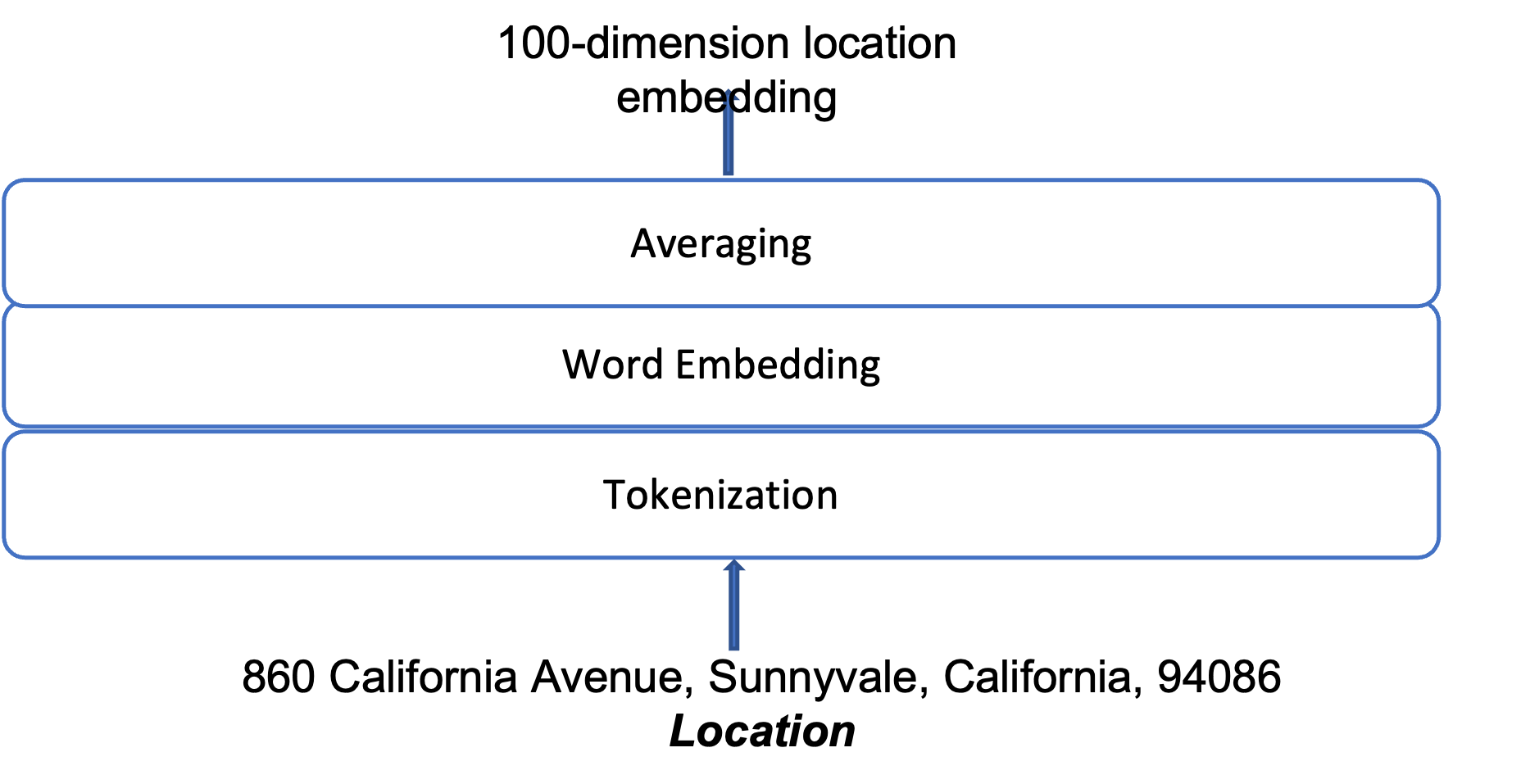}
	\caption{Location Embedding using Word Embedding.} \label{word}
\end{figure}

\begin{figure}[h]
	\centering
	\includegraphics[scale=0.5]{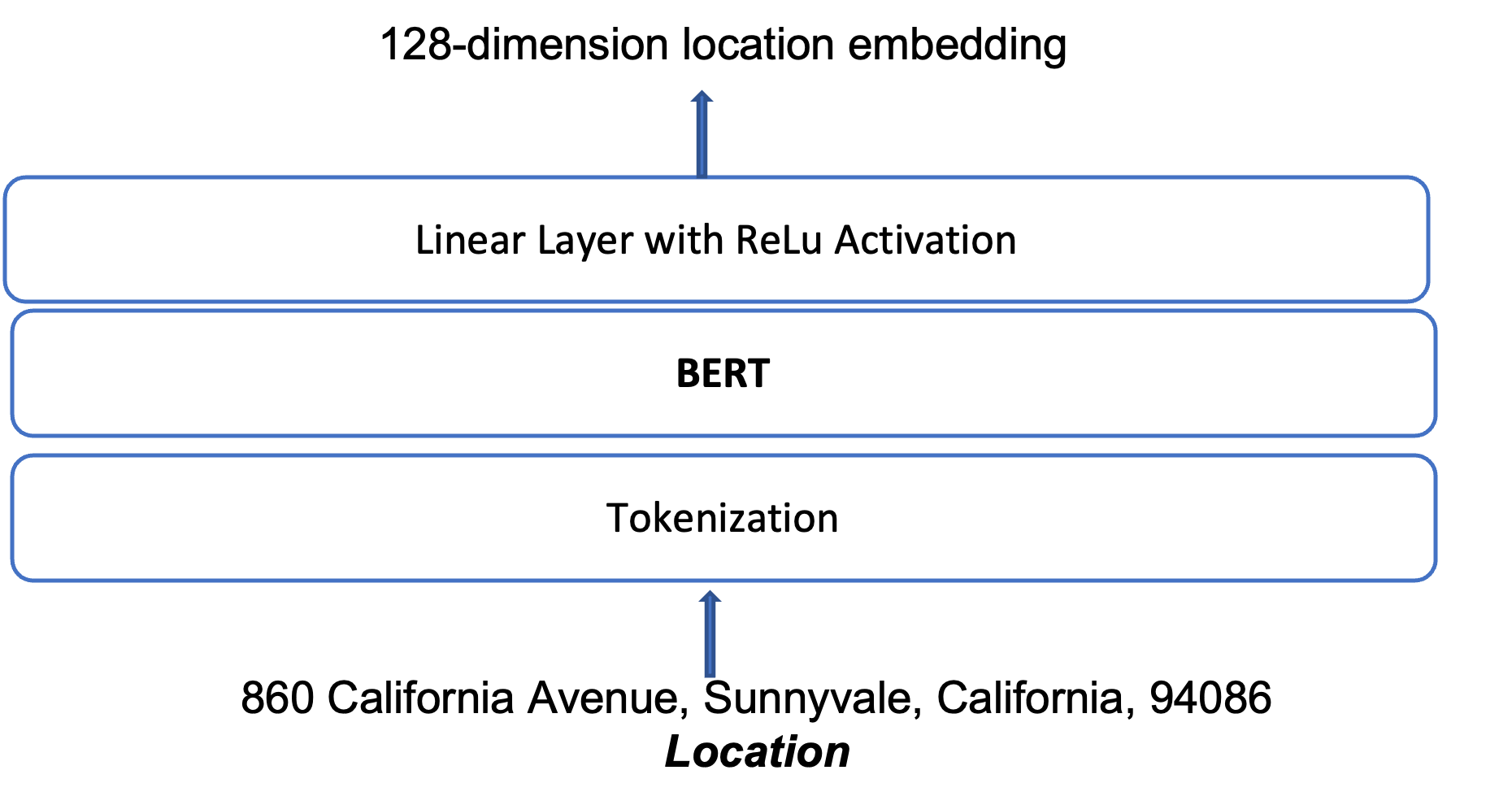}
	\caption{Location Embedding using BERT.} \label{bert}
\end{figure}

We tried a few different approaches to convert location text to location embedding.
One approach is to use a pre-trained word embedding model in PyTorch (GloVe), which is illustrated in Figure \ref{word}.
Another approach is to use the state-of-the-art BERT model \cite{bert2018} for text representation learning, which is shown in Figure \ref{bert}.
We also fine tune the pre-trained BERT model in our training process.

\section{Experimental Results}

For evaluation, we setup the training, validation and testing datasets by different time windows.
For example, we can use data of last $n$ years as training data, data of next one month or one quarter as validation and testing data.
The evaluation metric for model quality is mean absolute error (MAE).
Due to Walmart’s privacy policy, the results are presented as percentage proportions to the baseline embedding model.

In this section, we compare the quality and inference time of different model setups, from different numbers of fully connected layers to different location embedding methods.
The baseline model is using two fully connected ReLU layers and no location embedding.
The results are shown in Table \ref{quality} and Table \ref{time}.

\begin{table}[ht]
  \caption{Quality (MAE) of Embedding Model with Different Setups.}
  \label{quality}
  \small
  \begin{tabular}{|c|c|c|c|c|c|}
    \hline
     & No Location & Word Embedding & BERT \\
    \hline
    2 layers & 100\% & 97\% & 91\% \\ \hline
    3 layers & 94\% & 87\% & 86\%  \\   \hline
  \end{tabular}
\end{table}

\begin{table}[ht]
  \caption{Inference Time of Embedding Model with Different Setups.}
  \label{time}
  \small
  \begin{tabular}{|c|c|c|c|c|c|}
    \hline
     & No Location & Word Embedding & BERT \\
    \hline
    2 layers & 100\% & 110\% & >500\% \\ \hline
    3 layers & 105\% & 115\% & >500\%  \\   \hline
  \end{tabular}
\end{table}

In both Table \ref{quality} and Table \ref{time}, lower percentage indicates better model performance.
Although embedding model with BERT is the best one in terms of model quality, its inference latency increases by 4 times compared to baseline, which is very slow.
Considering we need to process hundreds of millions of customers offline, the total running time of our system using BERT is not ideal.
Embedding model with word embedding strikes a good balance between quality and inference latency.

\section{Conclusions}

In this paper, we propose an effective and efficient marketing audience
expansion system, which can handle hundreds of millions of customers.
We use a deep learning model to generate customer embeddings.
The deep learning model can handle both dense numerical features and sparse categorical features. 
And the model encodes location embedding using transfer learning
Our system has great adaptability by constructing different similarity metrics for different campaigns.
The similarity metrics are interpretable and meaningful and can reflect different business metrics.
Extensive experiments have been conducted to demonstrate the scalability, efficiency and quality of our system and embedding
model.

There are several directions for the future work in our marketing audience expansion system.
First, we can explore the direction of combining both similarity-based and classification-based approaches for searching lookalike customers.
Second, we can study how to filter and rank the lookalike customers using machine learning models to improve ranking quality.


\bibliographystyle{ACM-Reference-Format}
\bibliography{reference}


\begin{thebibliography}{34}


\ifx \showCODEN    \undefined \def \showCODEN     #1{\unskip}     \fi
\ifx \showDOI      \undefined \def \showDOI       #1{#1}\fi
\ifx \showISBNx    \undefined \def \showISBNx     #1{\unskip}     \fi
\ifx \showISBNxiii \undefined \def \showISBNxiii  #1{\unskip}     \fi
\ifx \showISSN     \undefined \def \showISSN      #1{\unskip}     \fi
\ifx \showLCCN     \undefined \def \showLCCN      #1{\unskip}     \fi
\ifx \shownote     \undefined \def \shownote      #1{#1}          \fi
\ifx \showarticletitle \undefined \def \showarticletitle #1{#1}   \fi
\ifx \showURL      \undefined \def \showURL       {\relax}        \fi
\providecommand\bibfield[2]{#2}
\providecommand\bibinfo[2]{#2}
\providecommand\natexlab[1]{#1}
\providecommand\showeprint[2][]{arXiv:#2}

\bibitem[\protect\citeauthoryear{Amirrahmat, Alshibli, Jarrar, Zhang, and
  Regueiro}{Amirrahmat et~al\mbox{.}}{2018}]%
        {amirrahmat2018}
\bibfield{author}{\bibinfo{person}{S Amirrahmat}, \bibinfo{person}{KA
  Alshibli}, \bibinfo{person}{MF Jarrar}, \bibinfo{person}{B Zhang}, {and}
  \bibinfo{person}{RA Regueiro}.} \bibinfo{year}{2018}\natexlab{}.
\newblock \showarticletitle{Equivalent continuum strain calculations based on
  3D particle kinematic measurements of sand}.
\newblock \bibinfo{journal}{\emph{International Journal for Numerical and
  Analytical Methods in Geomechanics}}  \bibinfo{volume}{42}
  (\bibinfo{year}{2018}), \bibinfo{pages}{999--1015}.
\newblock
Issue 8.
\urldef\tempurl%
\url{https://doi.org/10.1002/nag.2779}
\showDOI{\tempurl}


\bibitem[\protect\citeauthoryear{Devlin, Chang, Lee, and Toutanova}{Devlin
  et~al\mbox{.}}{2018}]%
        {bert2018}
\bibfield{author}{\bibinfo{person}{Jacob Devlin}, \bibinfo{person}{Ming-Wei
  Chang}, \bibinfo{person}{Kenton Lee}, {and} \bibinfo{person}{Kristina
  Toutanova}.} \bibinfo{year}{2018}\natexlab{}.
\newblock \bibinfo{title}{BERT: Pre-training of Deep Bidirectional Transformers
  for Language Understanding}.
\newblock
\newblock
\urldef\tempurl%
\url{https://doi.org/10.48550/ARXIV.1810.04805}
\showDOI{\tempurl}


\bibitem[\protect\citeauthoryear{deWet and Ou}{deWet and Ou}{2019}]%
        {dewet2019finding}
\bibfield{author}{\bibinfo{person}{Stephanie deWet} {and}
  \bibinfo{person}{Jiafan Ou}.} \bibinfo{year}{2019}\natexlab{}.
\newblock \showarticletitle{Finding users who act alike: transfer learning for
  expanding advertiser audiences}. In \bibinfo{booktitle}{\emph{Proceedings of
  the 25th ACM SIGKDD International Conference on Knowledge Discovery \& Data
  Mining}}. \bibinfo{pages}{2251--2259}.
\newblock


\bibitem[\protect\citeauthoryear{Gong, Wang, and Peng}{Gong
  et~al\mbox{.}}{2017}]%
        {acm2017web}
\bibfield{author}{\bibinfo{person}{Dihong Gong}, \bibinfo{person}{Daisy~Zhe
  Wang}, {and} \bibinfo{person}{Yang Peng}.} \bibinfo{year}{2017}\natexlab{}.
\newblock \showarticletitle{Multimodal Learning for Web Information
  Extraction}. In \bibinfo{booktitle}{\emph{Proceedings of the 25th ACM
  International Conference on Multimedia}} \emph{(\bibinfo{series}{MM '17})}.
  \bibinfo{publisher}{Association for Computing Machinery},
  \bibinfo{address}{New York, NY, USA}, \bibinfo{pages}{288–296}.
\newblock
\showISBNx{9781450349062}


\bibitem[\protect\citeauthoryear{Grbovic and Cheng}{Grbovic and Cheng}{2018}]%
        {grbovic2018real}
\bibfield{author}{\bibinfo{person}{Mihajlo Grbovic} {and}
  \bibinfo{person}{Haibin Cheng}.} \bibinfo{year}{2018}\natexlab{}.
\newblock \showarticletitle{Real-time personalization using embeddings for
  search ranking at airbnb}. In \bibinfo{booktitle}{\emph{Proceedings of the
  24th ACM SIGKDD International Conference on Knowledge Discovery \& Data
  Mining}}. \bibinfo{pages}{311--320}.
\newblock


\bibitem[\protect\citeauthoryear{Guo, Sun, Lindgren, Geng, Simcha, Chern, and
  Kumar}{Guo et~al\mbox{.}}{2020}]%
        {scann2020}
\bibfield{author}{\bibinfo{person}{Ruiqi Guo}, \bibinfo{person}{Philip Sun},
  \bibinfo{person}{Erik Lindgren}, \bibinfo{person}{Quan Geng},
  \bibinfo{person}{David Simcha}, \bibinfo{person}{Felix Chern}, {and}
  \bibinfo{person}{Sanjiv Kumar}.} \bibinfo{year}{2020}\natexlab{}.
\newblock \showarticletitle{Accelerating Large-Scale Inference with Anisotropic
  Vector Quantization}. In \bibinfo{booktitle}{\emph{International Conference
  on Machine Learning}}.
\newblock
\urldef\tempurl%
\url{https://arxiv.org/abs/1908.10396}
\showURL{%
\tempurl}


\bibitem[\protect\citeauthoryear{Jarrow, Murataj, Wells, and Zhu}{Jarrow
  et~al\mbox{.}}{2021}]%
        {jarrow2021low}
\bibfield{author}{\bibinfo{person}{Robert~A. Jarrow}, \bibinfo{person}{Rinald
  Murataj}, \bibinfo{person}{Martin~T. Wells}, {and} \bibinfo{person}{Liao
  Zhu}.} \bibinfo{year}{2021}\natexlab{}.
\newblock \showarticletitle{The Low-volatility Anomaly and the Adaptive
  Multi-Factor Model}.
\newblock \bibinfo{journal}{\emph{arXiv preprint arXiv:2003.08302}}
  (\bibinfo{year}{2021}).
\newblock


\bibitem[\protect\citeauthoryear{Jiang, Lin, Yao, and Lu}{Jiang
  et~al\mbox{.}}{2019}]%
        {jiang2019comprehensive}
\bibfield{author}{\bibinfo{person}{Jinling Jiang}, \bibinfo{person}{Xiaoming
  Lin}, \bibinfo{person}{Junjie Yao}, {and} \bibinfo{person}{Hua Lu}.}
  \bibinfo{year}{2019}\natexlab{}.
\newblock \showarticletitle{Comprehensive audience expansion based on
  end-to-end neural prediction}. In \bibinfo{booktitle}{\emph{CEUR Workshop
  Proceedings}}, Vol.~\bibinfo{volume}{2410}. CEUR Workshop Proceedings.
\newblock


\bibitem[\protect\citeauthoryear{Jie, Wang, Xu, and Shen}{Jie
  et~al\mbox{.}}{2022a}]%
        {jiemulti2022}
\bibfield{author}{\bibinfo{person}{Cheng Jie}, \bibinfo{person}{Zigeng Wang},
  \bibinfo{person}{Da Xu}, {and} \bibinfo{person}{Wei Shen}.}
  \bibinfo{year}{2022}\natexlab{a}.
\newblock \showarticletitle{Multi-objective Cluster Based Bidding Algorithm for
  E-Commerce Search Engine Marketing System}.
\newblock \bibinfo{journal}{\emph{Frontiers in Big Data}}
  (\bibinfo{year}{2022}), \bibinfo{pages}{77}.
\newblock


\bibitem[\protect\citeauthoryear{Jie, Xu, Wang, and Shen}{Jie
  et~al\mbox{.}}{2022b}]%
        {jiedeep2022}
\bibfield{author}{\bibinfo{person}{Cheng Jie}, \bibinfo{person}{Da Xu},
  \bibinfo{person}{Zigeng Wang}, {and} \bibinfo{person}{Wei Shen}.}
  \bibinfo{year}{2022}\natexlab{b}.
\newblock \bibinfo{title}{Deep Learning Based Page Creation for Improving
  E-Commerce Organic Search Traffic}.
\newblock
\newblock
\urldef\tempurl%
\url{https://doi.org/10.48550/ARXIV.2209.10792}
\showDOI{\tempurl}


\bibitem[\protect\citeauthoryear{Jie, Xu, Wang, Wang, and Shen}{Jie
  et~al\mbox{.}}{2021}]%
        {jie2021bidding}
\bibfield{author}{\bibinfo{person}{Cheng Jie}, \bibinfo{person}{Da Xu},
  \bibinfo{person}{Zigeng Wang}, \bibinfo{person}{Lu Wang}, {and}
  \bibinfo{person}{Wei Shen}.} \bibinfo{year}{2021}\natexlab{}.
\newblock \bibinfo{title}{An Efficient Group-based Search Engine Marketing
  System for E-Commerce}.
\newblock
\newblock
\urldef\tempurl%
\url{https://doi.org/10.48550/ARXIV.2106.12700}
\showDOI{\tempurl}


\bibitem[\protect\citeauthoryear{Johnson, Douze, and J{\'e}gou}{Johnson
  et~al\mbox{.}}{2019}]%
        {faiss2019}
\bibfield{author}{\bibinfo{person}{Jeff Johnson}, \bibinfo{person}{Matthijs
  Douze}, {and} \bibinfo{person}{Herv{\'e} J{\'e}gou}.}
  \bibinfo{year}{2019}\natexlab{}.
\newblock \showarticletitle{Billion-scale similarity search with {GPUs}}.
\newblock \bibinfo{journal}{\emph{IEEE Transactions on Big Data}}
  \bibinfo{volume}{7}, \bibinfo{number}{3} (\bibinfo{year}{2019}),
  \bibinfo{pages}{535--547}.
\newblock


\bibitem[\protect\citeauthoryear{Li, Song, Sun, and Zhu}{Li
  et~al\mbox{.}}{2021}]%
        {li2021frequentnet}
\bibfield{author}{\bibinfo{person}{Yifei Li}, \bibinfo{person}{Kuangyan Song},
  \bibinfo{person}{Yiming Sun}, {and} \bibinfo{person}{Liao Zhu}.}
  \bibinfo{year}{2021}\natexlab{}.
\newblock \showarticletitle{FrequentNet: A Novel Interpretable Deep Learning
  Model for Image Classification}.
\newblock \bibinfo{journal}{\emph{arXiv preprint arXiv:2001.01034}}
  (\bibinfo{year}{2021}).
\newblock


\bibitem[\protect\citeauthoryear{Liu, Pardoe, Liu, Thakur, Cao, and Li}{Liu
  et~al\mbox{.}}{2016}]%
        {liu2016audience}
\bibfield{author}{\bibinfo{person}{Haishan Liu}, \bibinfo{person}{David
  Pardoe}, \bibinfo{person}{Kun Liu}, \bibinfo{person}{Manoj Thakur},
  \bibinfo{person}{Frank Cao}, {and} \bibinfo{person}{Chongzhe Li}.}
  \bibinfo{year}{2016}\natexlab{}.
\newblock \showarticletitle{Audience expansion for online social network
  advertising}. In \bibinfo{booktitle}{\emph{Proceedings of the 22nd ACM SIGKDD
  International Conference on Knowledge Discovery and Data Mining}}.
  \bibinfo{pages}{165--174}.
\newblock


\bibitem[\protect\citeauthoryear{Luo, Chen, Zhang, Ren, Zhang, Regueiro,
  Alshibli, and Lu}{Luo et~al\mbox{.}}{2022}]%
        {luo2022constitutive}
\bibfield{author}{\bibinfo{person}{Huiyang Luo}, \bibinfo{person}{Huiluo Chen},
  \bibinfo{person}{Runyu Zhang}, \bibinfo{person}{Yao Ren},
  \bibinfo{person}{Boning Zhang}, \bibinfo{person}{Richard~A. Regueiro},
  \bibinfo{person}{Khalid Alshibli}, {and} \bibinfo{person}{Hongbing Lu}.}
  \bibinfo{year}{2022}\natexlab{}.
\newblock \showarticletitle{4 - Constitutive behavior of granular materials
  under high rate of uniaxial strain loading}.
\newblock In \bibinfo{booktitle}{\emph{Advances in Experimental Impact
  Mechanics}}, \bibfield{editor}{\bibinfo{person}{Bo~Song}} (Ed.).
  \bibinfo{publisher}{Elsevier}, \bibinfo{pages}{99--124}.
\newblock
\showISBNx{978-0-12-823325-2}
\urldef\tempurl%
\url{https://doi.org/10.1016/B978-0-12-823325-2.00005-4}
\showDOI{\tempurl}


\bibitem[\protect\citeauthoryear{Ma, Wen, Xia, and Chen}{Ma
  et~al\mbox{.}}{2016}]%
        {pmlrma16}
\bibfield{author}{\bibinfo{person}{Qiang Ma}, \bibinfo{person}{Musen Wen},
  \bibinfo{person}{Zhen Xia}, {and} \bibinfo{person}{Datong Chen}.}
  \bibinfo{year}{2016}\natexlab{}.
\newblock \showarticletitle{A Sub-linear, Massive-scale Look-alike Audience
  Extension System A Massive-scale Look-alike Audience Extension}. In
  \bibinfo{booktitle}{\emph{Proceedings of the 5th International Workshop on
  Big Data, Streams and Heterogeneous Source Mining: Algorithms, Systems,
  Programming Models and Applications at KDD 2016}}
  \emph{(\bibinfo{series}{Proceedings of Machine Learning Research})},
  \bibfield{editor}{\bibinfo{person}{Wei Fan}, \bibinfo{person}{Albert Bifet},
  \bibinfo{person}{Jesse Read}, \bibinfo{person}{Qiang Yang}, {and}
  \bibinfo{person}{Philip~S. Yu}} (Eds.), Vol.~\bibinfo{volume}{53}.
  \bibinfo{publisher}{PMLR}, \bibinfo{address}{San Francisco, California, USA},
  \bibinfo{pages}{51--67}.
\newblock
\urldef\tempurl%
\url{https://proceedings.mlr.press/v53/ma16.html}
\showURL{%
\tempurl}


\bibitem[\protect\citeauthoryear{Nia, Grant, Peng, Wang, and Petrovic}{Nia
  et~al\mbox{.}}{2013}]%
        {niauniversity}
\bibfield{author}{\bibinfo{person}{Morteza~Shahriari Nia},
  \bibinfo{person}{Christan Grant}, \bibinfo{person}{Yang Peng},
  \bibinfo{person}{Daisy~Zhe Wang}, {and} \bibinfo{person}{Milenko Petrovic}.}
  \bibinfo{year}{2013}\natexlab{}.
\newblock \showarticletitle{University of Florida Knowledge Base Acceleration
  Notebook}.
\newblock \bibinfo{journal}{\emph{The Twenty-Second Text REtrieval Conference
  (TREC 2013)}} (\bibinfo{year}{2013}).
\newblock


\bibitem[\protect\citeauthoryear{Nia, Grant, Peng, Wang, and Petrovic}{Nia
  et~al\mbox{.}}{2014}]%
        {nia2014streaming}
\bibfield{author}{\bibinfo{person}{Morteza~Shahriari Nia},
  \bibinfo{person}{Christan~Earl Grant}, \bibinfo{person}{Yang Peng},
  \bibinfo{person}{Daisy~Zhe Wang}, {and} \bibinfo{person}{Milenko Petrovic}.}
  \bibinfo{year}{2014}\natexlab{}.
\newblock \showarticletitle{Streaming Fact Extraction for Wikipedia Entities at
  Web-Scale.}. In \bibinfo{booktitle}{\emph{FLAIRS Conference}}.
\newblock


\bibitem[\protect\citeauthoryear{Pal, Eksombatchai, Zhou, Zhao, Rosenberg, and
  Leskovec}{Pal et~al\mbox{.}}{2020}]%
        {pal2020pinnersage}
\bibfield{author}{\bibinfo{person}{Aditya Pal}, \bibinfo{person}{Chantat
  Eksombatchai}, \bibinfo{person}{Yitong Zhou}, \bibinfo{person}{Bo Zhao},
  \bibinfo{person}{Charles Rosenberg}, {and} \bibinfo{person}{Jure Leskovec}.}
  \bibinfo{year}{2020}\natexlab{}.
\newblock \showarticletitle{Pinnersage: Multi-modal user embedding framework
  for recommendations at pinterest}. In \bibinfo{booktitle}{\emph{Proceedings
  of the 26th ACM SIGKDD International Conference on Knowledge Discovery \&
  Data Mining}}. \bibinfo{pages}{2311--2320}.
\newblock


\bibitem[\protect\citeauthoryear{Peng}{Peng}{2017}]%
        {peng2017multimodal}
\bibfield{author}{\bibinfo{person}{Yang Peng}.}
  \bibinfo{year}{2017}\natexlab{}.
\newblock \showarticletitle{Multimodal Fusion: A Theory and Applications}.
\newblock \bibinfo{journal}{\emph{University of Florida}}
  (\bibinfo{year}{2017}).
\newblock


\bibitem[\protect\citeauthoryear{Peng}{Peng}{2023}]%
        {peng2023web}
\bibfield{author}{\bibinfo{person}{Yang Peng}.}
  \bibinfo{year}{2023}\natexlab{}.
\newblock \showarticletitle{Query-Driven Knowledge Graph Construction Using
  Question Answering and Multimodal Fusion}. In
  \bibinfo{booktitle}{\emph{Companion Proceedings of the ACM Web Conference
  2023}} \emph{(\bibinfo{series}{WWW '23 Companion})}.
  \bibinfo{publisher}{Association for Computing Machinery},
  \bibinfo{address}{New York, NY, USA}, \bibinfo{pages}{1119–1126}.
\newblock
\showISBNx{9781450394192}
\urldef\tempurl%
\url{https://doi.org/10.1145/3543873.3587567}
\showDOI{\tempurl}


\bibitem[\protect\citeauthoryear{Peng and Wang}{Peng and Wang}{2022a}]%
        {peng2022qa}
\bibfield{author}{\bibinfo{person}{Yang Peng} {and} \bibinfo{person}{Daisy~Zhe
  Wang}.} \bibinfo{year}{2022}\natexlab{a}.
\newblock \bibinfo{title}{Knowledge Base Completion using Web-Based Question
  Answering and Multimodal Fusion}.
\newblock
\newblock
\urldef\tempurl%
\url{https://doi.org/10.48550/ARXIV.2211.07098}
\showDOI{\tempurl}


\bibitem[\protect\citeauthoryear{Peng and Wang}{Peng and Wang}{2022b}]%
        {peng2022kbc}
\bibfield{author}{\bibinfo{person}{Yang Peng} {and} \bibinfo{person}{Daisy~Zhe
  Wang}.} \bibinfo{year}{2022}\natexlab{b}.
\newblock \bibinfo{title}{Query-Driven Knowledge Base Completion using
  Multimodal Path Fusion over Multimodal Knowledge Graph}.
\newblock
\newblock
\urldef\tempurl%
\url{https://doi.org/10.48550/ARXIV.2212.01923}
\showDOI{\tempurl}


\bibitem[\protect\citeauthoryear{Peng, Wang, Patwa, Gong, and Fang}{Peng
  et~al\mbox{.}}{2015}]%
        {peng2015probabilistic}
\bibfield{author}{\bibinfo{person}{Yang Peng}, \bibinfo{person}{Daisy~Zhe
  Wang}, \bibinfo{person}{Ishan Patwa}, \bibinfo{person}{Dihong Gong}, {and}
  \bibinfo{person}{Chunsheng~Victor Fang}.} \bibinfo{year}{2015}\natexlab{}.
\newblock \showarticletitle{Probabilistic Ensemble Fusion for Multimodal Word
  Sense Disambiguation}. In \bibinfo{booktitle}{\emph{Multimedia (ISM), 2015
  IEEE International Symposium on}}. IEEE, \bibinfo{pages}{172--177}.
\newblock


\bibitem[\protect\citeauthoryear{Peng, Zhou, Wang, and Fang}{Peng
  et~al\mbox{.}}{2016a}]%
        {peng2016scalable}
\bibfield{author}{\bibinfo{person}{Yang Peng}, \bibinfo{person}{Xiaofeng Zhou},
  \bibinfo{person}{Daisy~Zhe Wang}, {and} \bibinfo{person}{Chunsheng~Victor
  Fang}.} \bibinfo{year}{2016}\natexlab{a}.
\newblock \showarticletitle{Scalable image retrieval with multimodal fusion}.
  In \bibinfo{booktitle}{\emph{The Twenty-Ninth International Flairs
  Conference}}.
\newblock


\bibitem[\protect\citeauthoryear{Peng, Zhou, Wang, Patwa, Gong, and Fang}{Peng
  et~al\mbox{.}}{2016b}]%
        {peng2016multimodal}
\bibfield{author}{\bibinfo{person}{Yang Peng}, \bibinfo{person}{Xiaofeng Zhou},
  \bibinfo{person}{Daisy~Zhe Wang}, \bibinfo{person}{Ishan Patwa},
  \bibinfo{person}{Dihong Gong}, {and} \bibinfo{person}{Chunsheng Fang}.}
  \bibinfo{year}{2016}\natexlab{b}.
\newblock \showarticletitle{Multimodal Ensemble Fusion for Disambiguation and
  Retrieval}.
\newblock \bibinfo{journal}{\emph{IEEE MultiMedia}} (\bibinfo{year}{2016}).
\newblock


\bibitem[\protect\citeauthoryear{Senseney, Duan, Zhang, and Regueiro}{Senseney
  et~al\mbox{.}}{2017}]%
        {Senseney2017}
\bibfield{author}{\bibinfo{person}{Christopher~T Senseney},
  \bibinfo{person}{Zheng Duan}, \bibinfo{person}{Boning Zhang}, {and}
  \bibinfo{person}{Richard~A Regueiro}.} \bibinfo{year}{2017}\natexlab{}.
\newblock \showarticletitle{Combined spheropolyhedral discrete element
  (DE)-finite element (FE) computational modeling of vertical plate loading on
  cohesionless soil}.
\newblock \bibinfo{journal}{\emph{Acta Geotechnica}}  \bibinfo{volume}{12}
  (\bibinfo{year}{2017}), \bibinfo{pages}{593--603}.
\newblock
\urldef\tempurl%
\url{https://doi.org/10.1007/s11440-016-0519-8}
\showDOI{\tempurl}


\bibitem[\protect\citeauthoryear{Zhang, Herbold, Homel, and Regueiro}{Zhang
  et~al\mbox{.}}{2015}]%
        {Zhang2015fracture}
\bibfield{author}{\bibinfo{person}{Boning Zhang}, \bibinfo{person}{Eric~B.
  Herbold}, \bibinfo{person}{Michael~A. Homel}, {and}
  \bibinfo{person}{Richard~A. Regueiro}.} \bibinfo{year}{2015}\natexlab{}.
\newblock \bibinfo{booktitle}{\emph{{DEM Particle Fracture Model}}}.
\newblock \bibinfo{type}{{T}echnical {R}eport}. \bibinfo{institution}{Lawrence
  Livermore National Lab.(LLNL), Livermore, CA (United States)}.
\newblock


\bibitem[\protect\citeauthoryear{Zhang, Regueiro, Druckrey, and Alshibli}{Zhang
  et~al\mbox{.}}{2018}]%
        {zhang2016construction}
\bibfield{author}{\bibinfo{person}{Boning Zhang}, \bibinfo{person}{Richard~A.
  Regueiro}, \bibinfo{person}{Andrew~M. Druckrey}, {and}
  \bibinfo{person}{Khalid Alshibli}.} \bibinfo{year}{2018}\natexlab{}.
\newblock \showarticletitle{Construction of poly-ellipsoidal grain shapes from
  SMT imaging on sand, and the development of a new DEM contact detection
  algorithm}.
\newblock \bibinfo{journal}{\emph{Engineering Computations}}
  \bibinfo{volume}{35}, \bibinfo{number}{2} (\bibinfo{year}{2018}),
  \bibinfo{pages}{733--771}.
\newblock
\urldef\tempurl%
\url{https://doi.org/10.1108/EC-01-2017-0026}
\showDOI{\tempurl}


\bibitem[\protect\citeauthoryear{Zhu}{Zhu}{2020}]%
        {zhu2020adaptive}
\bibfield{author}{\bibinfo{person}{Liao Zhu}.} \bibinfo{year}{2020}\natexlab{}.
\newblock \bibinfo{booktitle}{\emph{The Adaptive Multi-Factor Model and the
  Financial Market}}.
\newblock \bibinfo{publisher}{eCommons}.
\newblock


\bibitem[\protect\citeauthoryear{Zhu, Basu, Jarrow, and Wells}{Zhu
  et~al\mbox{.}}{2020}]%
        {zhu2020high}
\bibfield{author}{\bibinfo{person}{Liao Zhu}, \bibinfo{person}{Sumanta Basu},
  \bibinfo{person}{Robert~A. Jarrow}, {and} \bibinfo{person}{Martin~T. Wells}.}
  \bibinfo{year}{2020}\natexlab{}.
\newblock \showarticletitle{High-Dimensional Estimation, Basis Assets, and the
  Adaptive Multi-Factor Model}.
\newblock \bibinfo{journal}{\emph{The Quarterly Journal of Finance}}
  \bibinfo{volume}{10}, \bibinfo{number}{04} (\bibinfo{year}{2020}),
  \bibinfo{pages}{2050017}.
\newblock


\bibitem[\protect\citeauthoryear{Zhu, Jarrow, and Wells}{Zhu
  et~al\mbox{.}}{2021a}]%
        {zhu2021time}
\bibfield{author}{\bibinfo{person}{Liao Zhu}, \bibinfo{person}{Robert~A.
  Jarrow}, {and} \bibinfo{person}{Martin~T. Wells}.}
  \bibinfo{year}{2021}\natexlab{a}.
\newblock \showarticletitle{Time-Invariance Coefficients Tests with the
  Adaptive Multi-Factor Model}.
\newblock \bibinfo{journal}{\emph{The Quarterly Journal of Finance}}
  \bibinfo{volume}{11}, \bibinfo{number}{04} (\bibinfo{year}{2021}),
  \bibinfo{pages}{2150019}.
\newblock


\bibitem[\protect\citeauthoryear{Zhu, Sun, and Wells}{Zhu
  et~al\mbox{.}}{2021b}]%
        {zhu2021clustering}
\bibfield{author}{\bibinfo{person}{Liao Zhu}, \bibinfo{person}{Ningning Sun},
  {and} \bibinfo{person}{Martin~T. Wells}.} \bibinfo{year}{2021}\natexlab{b}.
\newblock \showarticletitle{Clustering Structure of Microstructure Measures}.
\newblock \bibinfo{journal}{\emph{arXiv preprint arXiv:2107.02283}}
  (\bibinfo{year}{2021}).
\newblock


\bibitem[\protect\citeauthoryear{Zhu, Wu, and Wells}{Zhu
  et~al\mbox{.}}{2021c}]%
        {zhu2021news}
\bibfield{author}{\bibinfo{person}{Liao Zhu}, \bibinfo{person}{Haoxuan Wu},
  {and} \bibinfo{person}{Martin~T. Wells}.} \bibinfo{year}{2021}\natexlab{c}.
\newblock \showarticletitle{A News-based Machine Learning Model for Adaptive
  Asset Pricing}.
\newblock \bibinfo{journal}{\emph{arXiv preprint arXiv:2106.07103}}
  (\bibinfo{year}{2021}).
\newblock


\end{thebibliography}

\end{document}